\documentclass{vgtc}

\usepackage{times}
\usepackage{enumitem}

\onlineid{1011}

\vgtccategory{Research}

\vgtcinsertpkg

\usepackage[utf8]{inputenc}
\usepackage[english]{babel}
\usepackage{amsmath}
\usepackage{amsfonts}
\usepackage{amssymb}
\usepackage{graphicx}
\usepackage{bm} 

\usepackage[dvipsnames]{xcolor}

\usepackage{caption}
\usepackage{subfig}

\usepackage[numbers]{natbib} 
\usepackage[hidelinks]{hyperref} 

\usepackage{wrapfig} 

\usepackage{float}
\floatstyle{plaintop}
\restylefloat{table}

\usepackage{natbib}

\usepackage{lineno}
\modulolinenumbers[5]

\title{Efficient Pose Selection for Interactive Camera Calibration}
\author{Pavel Rojtberg \thanks{pavel.rojtberg@igd.fraunhofer.de} \\
\parbox{1.4in}{\scriptsize \centering Fraunhofer IGD, Darmstadt \\ TU Darmstadt}
\and
Arjan Kuijper \thanks{arjan.kuijper@igd.fraunhofer.de} \\
\parbox{1.4in}{\scriptsize \centering Fraunhofer IGD, Darmstadt \\ TU Darmstadt}
}

\makeatletter

\hypersetup{
pdftitle={\@title}
}

\abstract{ 
The choice of poses for camera calibration with planar patterns is only rarely considered --- yet the calibration precision heavily depends on it.
This work presents a pose selection method that finds a compact and robust set of calibration poses and is suitable for interactive calibration.
Consequently, singular poses that would lead to an unreliable solution are avoided explicitly, while poses reducing the uncertainty of the calibration are favoured.
For this, we use uncertainty propagation.

Our method takes advantage of a self-identifying calibration pattern to track the camera pose in real-time.
This allows to iteratively guide the user to the target poses, until the desired quality level is reached. Therefore, only a sparse set of key-frames is needed for calibration.

The method is evaluated on separate training and testing sets, as well as on synthetic data. Our approach performs better than comparable solutions while requiring 30\% less calibration frames.}

\CCScatlist{ 
  \CCScat{I.2.10}{ARTIFICIAL INTELLIGENCE}{Vision and Scene Understanding}{Modeling and recovery of physical attributes};
	\CCScat{I.5.5}{PATTERN RECOGNITION}{Implementation}{Interactive systems};
}

\begin{document}

\firstsection{Introduction}

\maketitle

Camera calibration in the context of 3D computer vision is the process of determining the internal camera geometric and optical characteristics (intrinsic parameters) and optionally the position and orientation of the camera frame in the world coordinate system (extrinsic parameters) \cite{tsai1987versatile}.
The performance of many 3D computer vision algorithms directly depends on the quality of this calibration. Furthermore, calibration is a recurring task that has to be performed each time the setup is changed. Even if a camera is replaced by an equivalent from the same series, the intrinsic parameters may vary due to build inaccuracies. 
The prevalent approach to camera calibration \cite{zhang2000flexible} is based on acquiring multiple images of a planar pattern of known size.

However, there are degenerate pose configurations \cite{sturm1999plane} that lead to unreliable solutions. Therefore, the task of calibration cannot be performed by inexperienced users --- even researchers working in the field often struggle to quantify what constitutes good calibration images.

There has been research on the effect of the angle between image plane and pattern on the estimation error; \citet{triggs1998autocalibration} related the angular spread to the error in focal length. He found a spread of more than $5^\circ$ necessary.
\citet{sturm1999plane} further differentiated between estimating principal point and focal length. More importantly, they discussed possible singularities when using one and two planes for calibration and related them to the individual pinhole parameters; e.g. if the pattern is parallel to the image plane in every frame, the focal length cannot be determined.
These findings were replicated in \cite{zhang2000flexible}.
However, the effect of poses on the estimation of the distortion parameters or general camera-to-board poses have not been considered so far.

Another aspect is the quality and quantity of calibration data. \citet{sun2005requirements} evaluated the sensitivity of camera models to noise, training data quantity and the calibration accuracy in respect to model complexity. However, they only measured the residual error on the respective training set, which is subject to over-fitting.
To overcome this, \citet{richardson2013iros} introduce the Max Expected Reprojection Error (Max ERE) metric that instead correlates with the testing error and thus allows a meaningful test for convergence.
\begin{figure}
\includegraphics[width=0.24\textwidth]{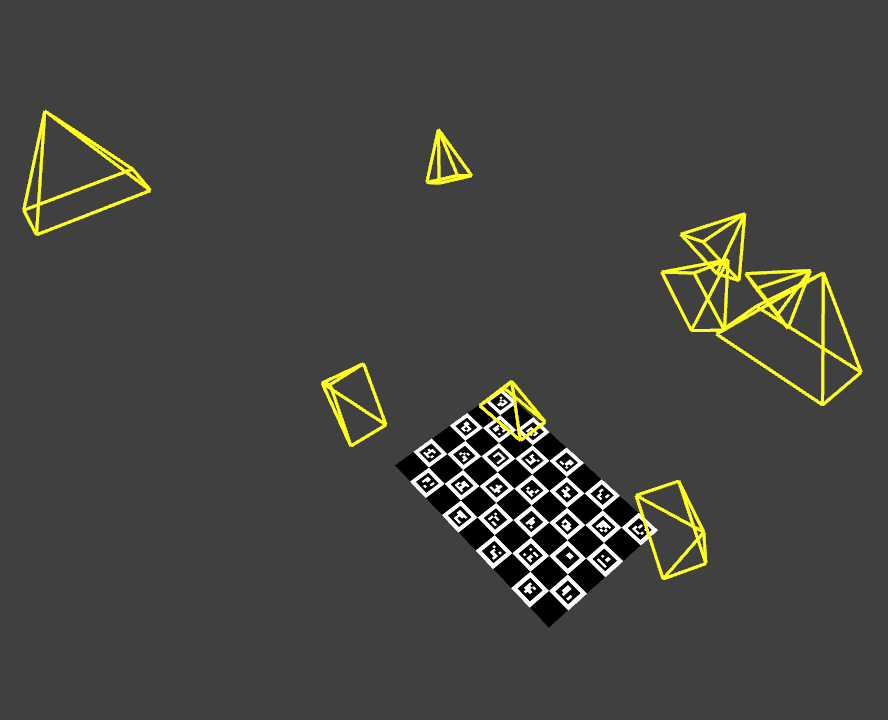}
\includegraphics[width=0.24\textwidth]{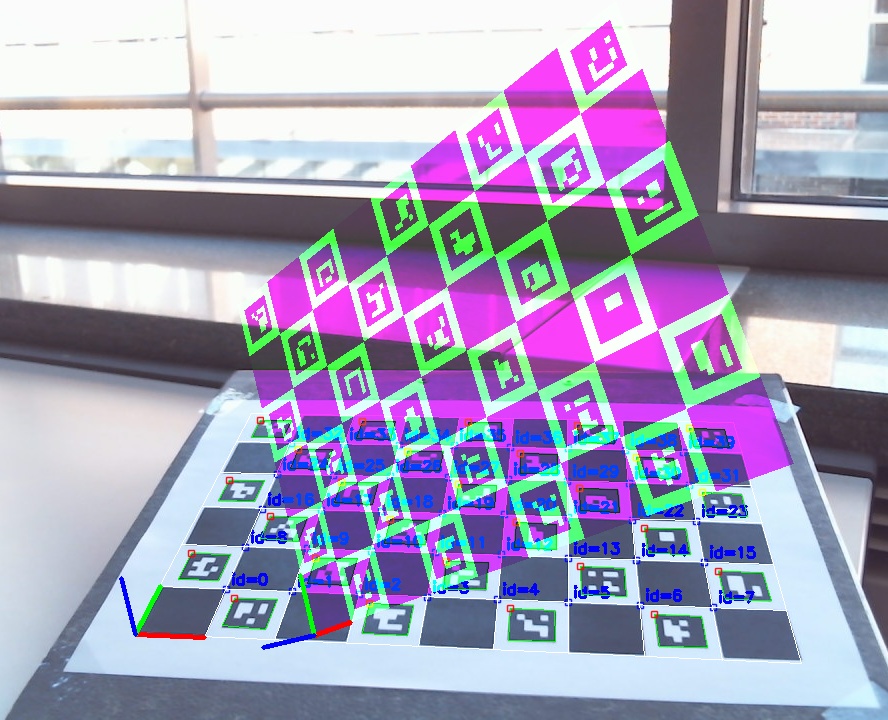}
\caption{Exemplary calibration using 9 selected poses and the user guidance overlay, projecting for the bottom right camera.}
\label{fig:init_overlay}
\end{figure}
Furthermore, they automatically compute a "best next pose" and use it for user guidance as an overlaid projection of the pattern. The poses are selected by performing an exhaustive search in a fixed set of about 60 candidate poses. For each pose a hypothetical calibration including this pose is performed and the pose that minimizes the Max ERE is selected.
However the candidate poses are uniformly distributed in the field of view and do not explicitly consider the angular spread and degenerate cases \cite{sturm1999plane}.

In the general context of user assistance for calibration tasks \cite{pankratz2015poster}, camera calibration was not yet specifically considered.

We propose to analytically generate optimal pattern poses while explicitly avoiding the degenerate pose configurations. For this we relate poses to constraints on individual parameters, such that the resulting pose sequence constrains all calibration parameters and ensures an accurate calibration.
This reduces the computation time from seconds to milliseconds compared to the exhaustive search of \cite{richardson2013iros}.

The uncertainty of the calibration parameters is assessed using the covariance of the estimated solution. The pose sequence is then adapted such that more constraints are captured for the most uncertain parameter.
The parameters covariance correlates with the testing error and therefore also serves as a convergence criterion.

Based on the above, our key contributions are;
\begin{enumerate}[nolistsep]
\item Empirical evidence for the need of two distinct pose selection strategies and
\item an efficient pose selection scheme for implementing both of them.
\end{enumerate}

This paper is structured as follows: in Section \ref{sec:immodel} the used camera model and the uncertainty estimation method are introduced and the choice of a suitable calibration pattern is discussed.
Section \ref{sec:poses} motivates and describes our novel pose selection method, while Section \ref{sec:process} describes the full calibration pipeline.
In Section \ref{sec:evaluation} the method is evaluated on real and synthetic data and compared with OpenCV \cite{opencv_library} and AprilCal \cite{richardson2013iros} calibration methods.
Furthermore, the compactness of the resulting calibration is analyzed and an informal user survey is performed to show the usability of the method.

We conclude with Section \ref{sec:conclusion} giving a summary of our results and discussing the limitations and future work.


\section{Preliminaries}
\label{sec:immodel}
We will use the pinhole camera model that, given the camera orientation $\mathbf{R}$, position $\mathbf{t}$ and the parameter vector $\mathbf{C}$, maps a 3D world point $ \mathbf{P} = [X, Y, Z] $ to a 2D image point $\mathbf{p} = [x, y]$:
\begin{equation}
\pi \left( \mathbf{P}; \mathbf{R}, \mathbf{t}, \mathbf{C} \right) = \mathbf{K} \, \Delta(\frac{1}{Z_c} \left[\mathbf{R} \; \mathbf{t} \right] \mathbf{P}).
\label{eq:cvcam}
\end{equation}
Here $\left[\mathbf{R} \; \mathbf{t} \right]$ is a 3x4 affine transformation, $Z_c$ denotes the depth of $\mathbf{P}$ after affine transformation, and $\textbf{K}$ is the camera calibration matrix containing the focal lengths (and aspect ratio) $\left[ f_x, f_y \right]$ and the principal point $\left[c_x, c_y \right]$.
\citet{zhang2000flexible} also includes a skew parameter $\gamma$ --- however, for CCD cameras it is safe to assume $\gamma$ to be zero \cite{sun2005requirements, hartley2005multiple}.
$\Delta(\cdot)$ models the commonly used \cite{sun2005requirements} radial \eqref{eq:raddist} and tangential \eqref{eq:tangdist} lens distortions (following \cite{heikkila1997four}) as 

\begin{subequations}
\begin{align}
\Delta(\mathbf{p}) =&\;\mathbf{p} \left( 1 + k_1 r^2 + k_2 r^4 + k_3 r^6 \right) \label{eq:raddist} \\
  & +
 \begin{pmatrix}
 2p_1xy+p_2 \left( r^2 + 2x^2 \right) \\
 p_1 \left( r^2 + 2y^2 \right) + 2p_2xy
 \end{pmatrix}, \label{eq:tangdist}
\end{align}
\end{subequations}
where $ r = \sqrt{x^2 + y^2} $.

Therefore $\textbf{C} = \left[ f_x, f_y, c_x, c_y, k_1, k_2, k_3, p_1, p_2 \right]$.

\subsection{Estimation and error analysis}

Given $M$ images each containing $N$ point correspondences, the underlying calibration method \cite{zhang2000flexible} minimizes the geometric error

\begin{equation}
\epsilon_{res} = \sum_{i}^N \sum_{j}^M \parallel \mathbf{p}_{ij} - \pi \left( \mathbf{P}_i; \mathbf{R}_j, \mathbf{t}_j, \mathbf{C} \right) \parallel^2,
\label{eq:reperr}
\end{equation}
where $\mathbf{p}_{ij}$ is an observed (noisy) 2D point in image $j$ and $\mathbf{P}_i$ is the corresponding 3D object point. 

Eq.\ \eqref{eq:reperr} is also referred to as the reprojection error and often used to assess the quality of a calibration. Yet, it only measures the residual error and is subject to over-fitting. Particularly $\epsilon_{res} = 0$ if exactly $N = 10.5$ point correspondences are used \cite[§7.1]{hartley2005multiple}.

The actual objective for calibration however, is the estimation error $\epsilon_{est}$, i.e. the distance between the solution and the (unknown) ground truth.
 \citet{richardson2013iros} propose the Max ERE as an alternative metric that correlates with the estimation error and also has a similar value range (pixels). However, it requires sampling and re-projecting the current solution.
Yet for user guidance and monitoring of convergence only the relative error of the parameters is needed. Therefore, we directly use the variance $\bm{\sigma}_{C}^2 $ of the estimated parameters. Particularly, we use the index of dispersion (IOD) $\sigma^2_i / C_i$ to ensure comparability of the parameters among each other.

Given the covariance of the image points $\mathbf{\Sigma}_{p}$ the backward transport of covariance \cite[§5.2.3]{hartley2005multiple} is used to obtain 
\begin{align} 
\mathbf{\Sigma}_{v} &= \left( \mathbf{J}^T \mathbf{\Sigma}_{p}^{-1} \mathbf{J} \right)^{+} \label{eq:variance} \\
\mathbf{J} &= \delta \pi / \delta \mathbf{v} \nonumber
\end{align}
where $\mathbf{J}$ is the Jacobian matrix, $\mathbf{v} = [\mathbf{C}, \mathbf{R}_1, \mathbf{t}_1, \ldots ,\mathbf{R}_M, \mathbf{t}_M]$ is the vector of unknowns and $(\cdot)^{+}$ denotes the pseudo inverse. 
For simplicity and because of the lack of prior knowledge we assume a standard deviation of 1px in each coordinate direction for the image points thus $ \mathbf{\Sigma}_{p} = \mathbf{I} $.

The diagonal entries of $\mathbf{\Sigma}_{v}$ contain the variance of the estimated $\mathbf{C}$. $\mathbf{J}$ is already computed in Levenberg-Marquardt step of \cite{zhang2000flexible}.

\subsection{Calibration pattern}
\label{sec:pattern}
Our approach works with any planar calibration target e.g. the common chessboard and circle grid patterns. However, for interactive user guidance a fast board detection is crucial. Therefore, we use the self-identifying ChArUco \cite{charuco} pattern  as implemented in OpenCV. This saves the time consuming ordering of the detected rectangles to a canonical topology when compared to the classical chessboard.
However, one can alternatively use any of the recently developed self-identifying targets \cite{atcheson2010caltag, birdal2016x, fiala2008self} here.

The pattern size is set to 9x6 squares resulting in up to 40 measurements at the chessboard joints per captured frame. This allows to successfully complete the initialization even if not all markers are detected as discussed in section \ref{sec:heuristics}.

\section{Pose selection}
\label{sec:poses}
The core idea of our approach is to explicitly specify individual key-frames which are used for calibration using the method of \citet{zhang2000flexible}. 

In this section first the relation of intrinsic parameters and board poses is discussed to motivate our split of the parameter vector into pinhole and distortion parameters.
For each parameter group we then present our set of rules to generate an optimal pose while explicitly avoiding degenerate configurations.

\subsection{Splitting pinhole and distortion parameters}
\label{sec:singsample}
Looking at eq.\ \eqref{eq:cvcam} we see that both $\mathbf{K}$ and $\Delta(\cdot)$ are applied at post-projection and thus describe 2D-to-2D mappings. Therefore, one might consider estimating $\mathbf{C}$ just from one board pose that uniformly samples the image. However, as both intrinsic and extrinsic parameters are estimated simultaneously by \cite{zhang2000flexible}, ambiguities arise.

Assuming $ \mathbf{R} = \mathbf{I} $ and the distortion parameters to be zero, by multiplying out \eqref{eq:cvcam} we get
\begin{equation}
\mathbf{p} = \begin{bmatrix} \dfrac{f_x (X + t_x)}{Z + t_z} + c_x \\
  \dfrac{f_y (Y + t_y)}{Z + t_z} + c_y \end{bmatrix}\label{eq:ambiguity}
\end{equation}
for all pattern points $\mathbf{P}$. In this case there are two ambiguities between
\begin{enumerate}[nolistsep]
\item the focal length $f$ and the distance to camera $t_z$ and
\item the in-plane translation $[t_x, t_y]$ and principal point $[c_x, c_y]$.
\end{enumerate}
These ambiguities can be resolved by requiring the pattern to be tilted towards the image plane such that there is only one $\mathbf{t}$ that satisfies eq.\ \eqref{eq:cvcam} for all pattern points.


\begin{figure}
\subfloat[Estimated distortion map] {
\includegraphics[width=0.24\textwidth]{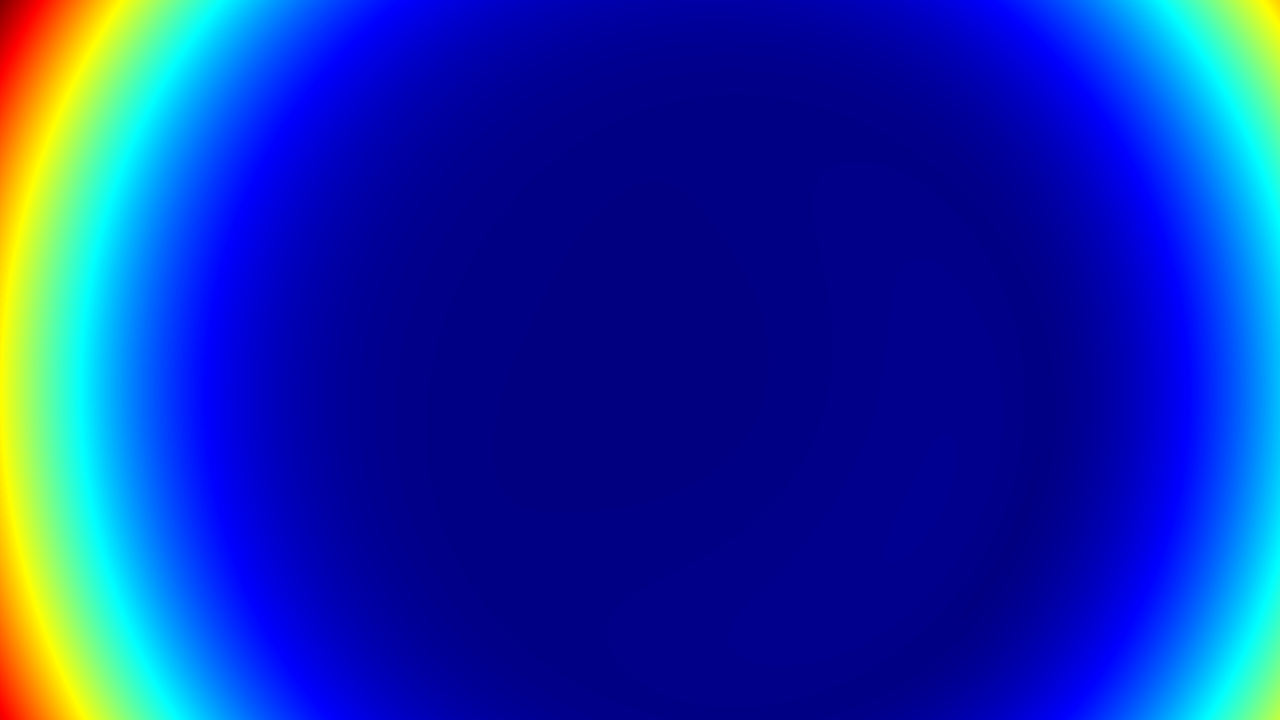}
\label{fig:dist_effect}
}
\subfloat[Target pose at max. distortion] {
\includegraphics[width=0.24\textwidth]{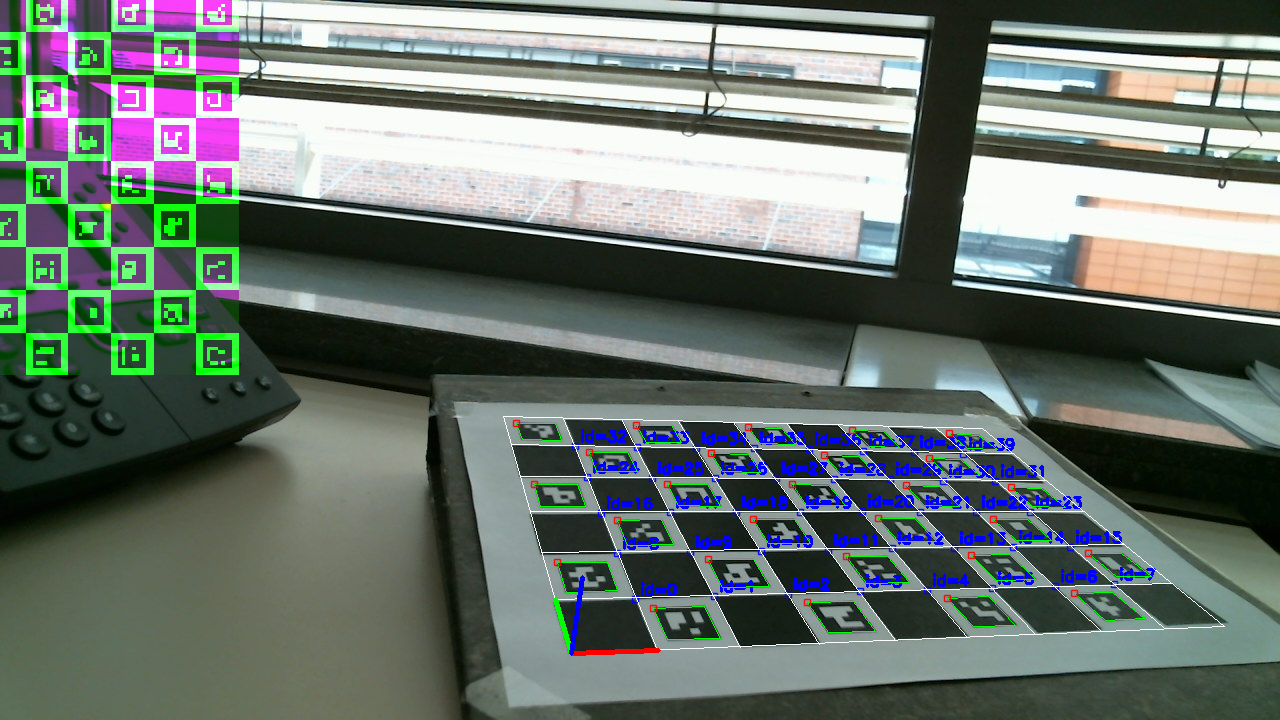}
}
\caption{Distortion map showing the magnitude of $\Delta(\textbf{p})$ for each pixel. To find the target pose we apply thresholding and fit an axis aligned bounding box.}
\label{fig:dist_pose}
\end{figure}

Considering the distortion parameters of $\Delta(\cdot)$ on the other hand, there are no similar ambiguities due to the non-linearity of the mapping. The parameters are rather determined by the maximal distortion strength evident in the image. Here it is more important to accurately measure the distortion in the corresponding image regions (see Figure \ref{fig:dist_effect}).

Therefore, we split the parameter vector $\textbf{C}$ into $\textbf{C}_K = \left[ f_x, f_y, c_x, c_y \right]$ and $\textbf{C}_\Delta = \left[ k_1, k_2, k_3, p_1, p_2 \right]$ and consider each group separately.

\subsection{Avoiding pinhole singularities}
While optimizing parameters in $\textbf{C}_K$, singular poses must be avoided. In addition to the case discussed above, we incorporate the cases identified in \cite{sturm1999plane}. Particularly, we restrict the 3D configuration of the calibration pattern as follows:
\begin{itemize}
\item The pattern must not be parallel to the image plane.
\item The pattern must not be parallel to one of the image axes.
\item Given two patterns, the "reflection constraint" must be fulfilled. This means that the vanishing lines of the two planes are not reflections of each other along both a horizontal and a vertical line in the image.
\end{itemize}
These restrictions ensure that each pose adds information that further constrains the pinhole parameters.

\subsection{Pose generation}
\label{sec:posegen}

\begin{figure}
\includegraphics[width=0.49\textwidth]{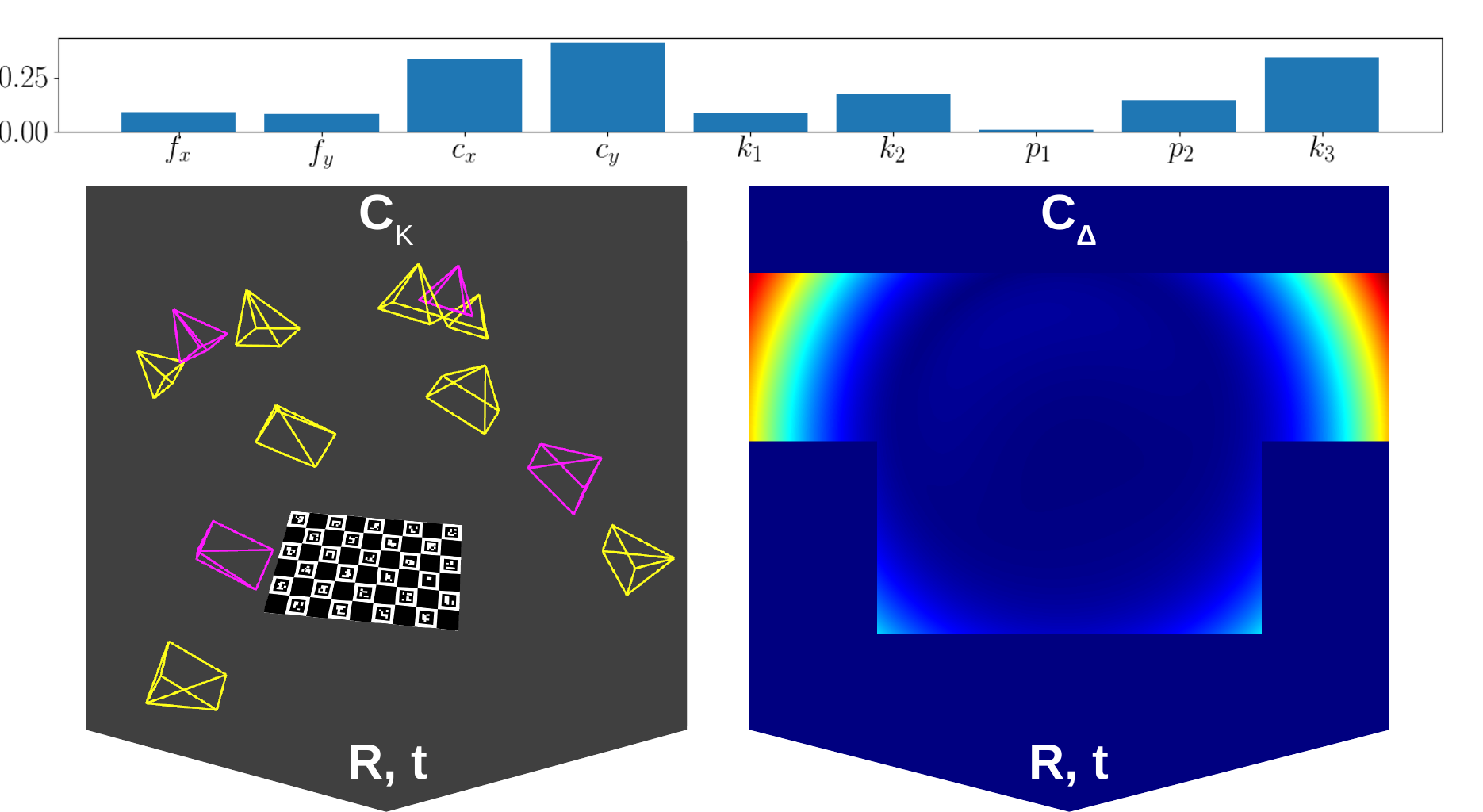}
\caption{Exemplary pose selection state. \textit{Top:} Index of dispersion. \textit{Left:} Intrinsic calibration position candidates after one (\textcolor{magenta}{magenta}) and two (\textcolor{Dandelion}{yellow}) subdivision steps . \textit{Right:} Distortion map with already visited regions masked out.}
\label{fig:posegen}
\end{figure}

As described in Section \ref{sec:singsample}, each parameter group requires a different strategy to generate an optimal calibration pose.

For the intrinsic parameters $\textbf{C}_K$ we follow \cite{triggs1998autocalibration, zhang2000flexible} and aim at  maximizing the angular spread between image plane and calibration pattern.
Accordingly, poses are generated as follows:
\begin{enumerate}
\item We choose a distance such that the whole pattern is visible, maximising the amount of observed 2D points.
\item Depending on the principal axis (e.g.\ $x$ for $f_y$) the pattern is tilted in the range of $(-70^\circ; 70^\circ)$ around that axis. The actual angle is interpolated using the sequence $\left[ 0.25, 0.75, 0.125, 0.375, \ldots \right]$ which corresponds to the binary subdivision of the $(0; 1)$ range (see Figure \ref{fig:posegen}). This strategy, as desired, maximizes the angular spread. 
\item The resulting pose would still be parallel to one of the image axes which prevents the estimation of the principal point along that axis \cite{sturm1999plane}. Therefore, the resulting view is rotated by $22.5^{\circ}$ which implements this requirement while keeping the principal orientation.
\item When determining $\left[c_x, c_y \right]$ the view is further shifted along the respective image axis by 5\% of the image size. This increases the spread along that axis and leads to faster convergence in our experiments.
\end{enumerate}

For the distortion parameters $\textbf{C}_\Delta$ the goal is to increase sampling accuracy in image regions exhibiting strong distortions. For this we generate a distortion map based on the current calibration estimate that encodes the displacement for each pixel. Using this map we search for the distorted regions as follows:

\begin{enumerate}
\item Threshold the distortion map (Figure \ref{fig:dist_effect}) to find the region with the strongest distortion.
\item Given the threshold image, an axis aligned bounding box (AABB) is fitted to the region, corresponding to a parallel view on the pattern. Note that the constraints for $\textbf{C}_K$ do not apply here.
\item The area covered by the AABB is excluded from subsequent searches (see Figure \ref{fig:posegen}). Effectively, the distorted regions are thereby visited in order of distortion strength.
\item The pattern is aligned with the top-left corner of the AABB and positioned at a depth s.t.\ its projection covers 33\% of the image width.
\end{enumerate}

The angular range and width limits mentioned above were set such that the calibration pattern could be reliably detected using the Logitech C525 camera.

\subsection{Initialization}
\label{sec:init}
The underlying calibration method \cite{zhang2000flexible} requires at least two views of the pattern for an initial solution which we select as follows:
\begin{itemize}
\item For the parameters $\textbf{C}_K$ a pose tilted by $45^{\circ}$ around $x$ is selected (see Section \ref{sec:posegen}). This particular angle was suggested by \cite{zhang2000flexible} and lies in between the extrema of $0^{\circ}$ where the focal length cannot be determined  and $90^{\circ}$ where the aspect ratio and principal point cannot be determined.
\item Without any prior knowledge we aim at an uniform sampling for estimating $\textbf{C}_\Delta$. To this end we compute a pose such that the pattern is parallel to the image plane and covers the whole view. While this violates the axis alignment requirements for $\textbf{C}_K$ poses, it still provides extra information as it is not coplanar to the first pose \cite{zhang2000flexible}. Furthermore, the reflection constraint is fulfilled.
\end{itemize}

To render an accurate overlay for the first pose without prior knowledge of the used camera, we employ a bootstraping strategy similar to \cite{richardson2013iros}; if the pattern can be detected, we perform a single frame calibration estimating the focal length only --- the principal point is fixed at the center and $\textbf{C}_\Delta$ is set to zero.

\section{Calibration process}
\label{sec:process}
In the following we present the parameter refinement and user guidance parts as well as any employed heuristics. This completes the calibration pipeline as used for the real data experiments.

\subsection{Parameter refinement}
\label{sec:param-refine}
After obtaining an initial solution using two key-frames, the goal is to minimize the cumulated variance $ \sum_i \sigma_i^2 \mid \sigma_i^2 \in \bm{\sigma}^2_C $ of the estimated parameters $\mathbf{C}$.
We approach this problem by targeting the variance of a single parameter $C_i \in \mathbf{C}$ at a time.
Here we pick the parameter with the highest index of dispersion (MaxIOD) $\sigma^2_i / C_i$ ($\sigma^2_i $ iff $ C_i = 0$).
Depending on the parameter group, a pose is then generated as described in Section \ref{sec:poses}.

For determining convergence, we use a ratio test of the parameter variance $r = \sigma^2_{i,n+1} / \sigma^2_{i,n}$. If the reduction $1 - r$ is below a given threshold, we assume the parameter to be converged and exclude it from further refinement. 
Here, we only consider parameters from the same group as there is typically only little reduction in the complementary group.
The calibration terminates once all parameters $\mathbf{C}$ have converged.

\subsection{User Guidance}
To guide the user, the targeted camera pose is projected using the current estimate of the intrinsic parameters. This projection is then displayed as an overlay on top of the live video stream (See Figure \ref{fig:init_overlay} and the video in the supplemental material).

To verify whether the user is sufficiently close to the target pose we use the Jaccard index $J(A, B)$ (intersection over union) computed from the area covered by the projection of pattern from the target pose $T$ and the area covered by the projection from the current pose estimate $E$. We assume that the user has reached the desired pose if $J(T, E) > 0.8$.

Comparing the projection overlap instead of using the estimated pose directly is more robust since the pose estimate is often unreliable --- especially during initialization.

\subsection{Heuristics}
\label{sec:heuristics}
Throughout the process we enforce the common heuristic \cite[§7.2]{hartley2005multiple} that the number of constraints should exceed the number of unknowns by a factor of five.
The used calibration method \cite{zhang2000flexible} not only estimates the intrinsic parameters $\mathbf{C}$, but also the relative pose of model plane and image plane i.e.\ the parameters $\mathbf{R}$, a 3D rotation, and $\mathbf{t}$, a 3D translation. When using $M$ calibration images we thus have $d = 9 + 6M$ unknowns and each point correspondence provides two constraints. For initialization ($M = 2$) we thus have $21$ unknowns, meaning $52.5$ point correspondences are needed in total or 27 correspondences per frame. For any subsequent frame only 15 points are required. 

To prevent inaccurate measurements due to motion blur and rolling shutter artifacts the pattern should be still. To ensure this we require all points to be re-detected in the consecutive frame and the mean motion of the points to be smaller then $1.5$px (determined empirically).




\begin{figure}
\subfloat[Poses 11-20 are optimizing $\textbf{C}_K$.] {
\includegraphics[width=0.49\textwidth]{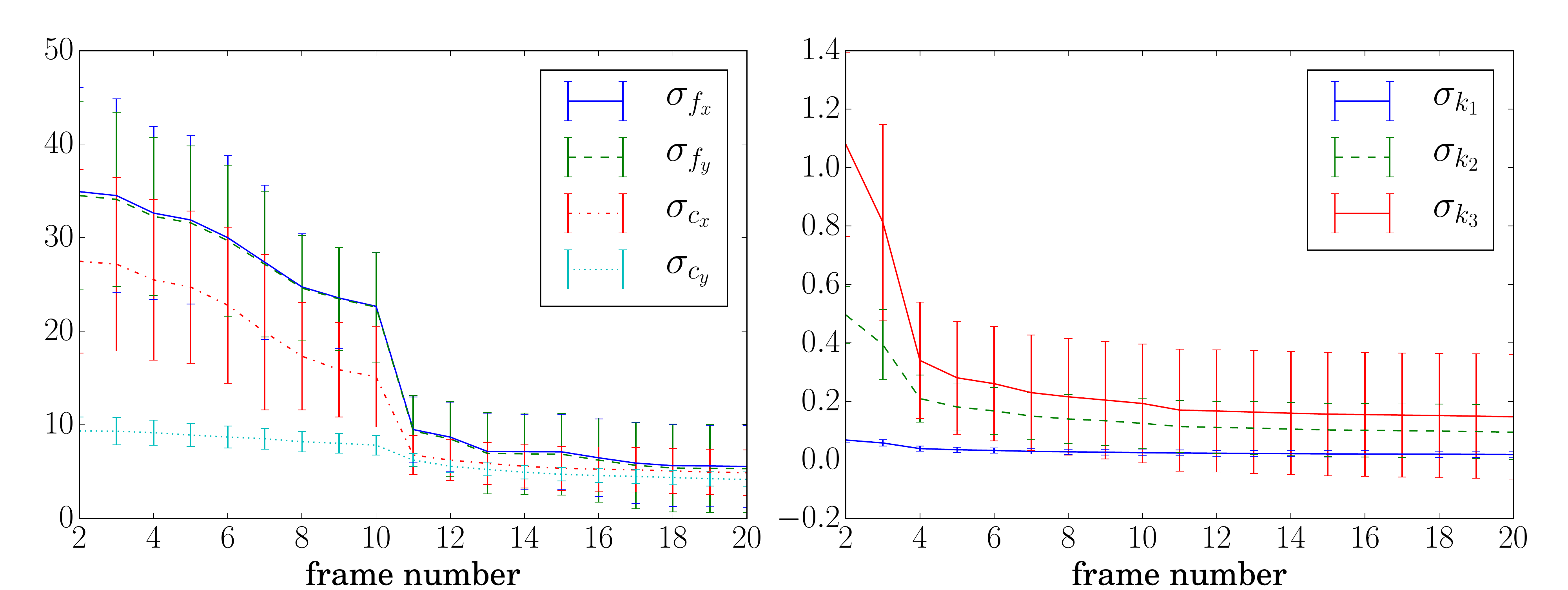}
\label{fig:std_kopt}
}
\\\hspace{\textwidth}
\subfloat[Poses 11-20 are optimizing $\textbf{C}_\Delta$.] {
\includegraphics[width=0.49\textwidth]{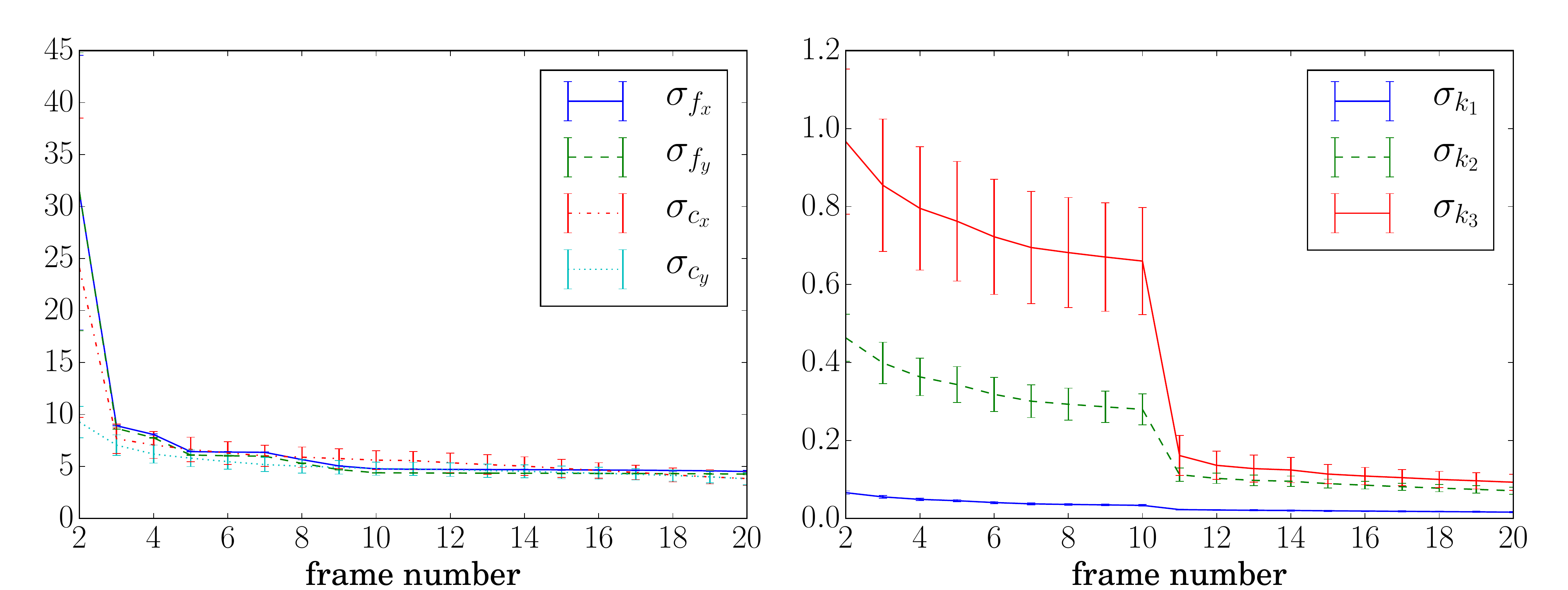}
\label{fig:std_distopt}
}
\caption{Correlation of pose selection strategies and calibration parameter uncertainty expressed using the standard deviation $\sigma$ (thus the error bars mean "variance of $\sigma$").\\
The first two poses are selected according to the initialization method. Poses 2-10 and 11-20 are selected by complementary strategies. Evaluated with synthetic images on 20 camera models sampled around the estimate of the Logitech C525 camera.}
\label{fig:correl}
\end{figure}

\section{Evaluation}
\label{sec:evaluation}

The presented method was evaluated on both synthetic and real data. The synthetic experiments aimed at validating the parameter splitting and pose generation rules presented in Section \ref{sec:poses}, while the real data was used for comparison with other methods.
Furthermore, the compactness of the results with real data was estimated by optimizing directly on the testing set.

\subsection{Synthetic data}
\label{sec:validation}
We performed multiple calibrations, each using 20 synthetic images. The first two camera poses were chosen as described in section \ref{sec:init} to allow a rough initial solution. The next 8 poses were chosen to optimize $\textbf{C}_\Delta$ while the last 10 poses were optimizing $\textbf{C}_K$ (and vice versa).

The camera parameters were based on the calibration parameters of a Logitech C525 camera $\textbf{C}_{real}$. However, the actual parameters were sampled around $\textbf{C}_{real}$ using a covariance matrix that allowed 10\% deviation for each of the parameters $\Sigma = diag(0.1 \cdot \textbf{C}_{real})$ as
\begin{equation}
\textbf{C} \sim \mathcal N(\textbf{C}_{real}, \Sigma).\label{eq:sample}
\end{equation}
Therefore, each synthetic calibration corresponds to using a different camera $\textbf{C}$ with known ground truth parameters. To allow generalization to different camera models, we kept the above pose generation sequence, but used 20 different cameras $\textbf{C}$.

Figure \ref{fig:correl} shows the mean standard deviation $\bm{\sigma}_C$ of the parameters. Notably there is a significant drop in $\sigma$ iff a pose matching the parameter group is used.

\begin{figure}
\subfloat[] {
\includegraphics[width=0.24\textwidth]{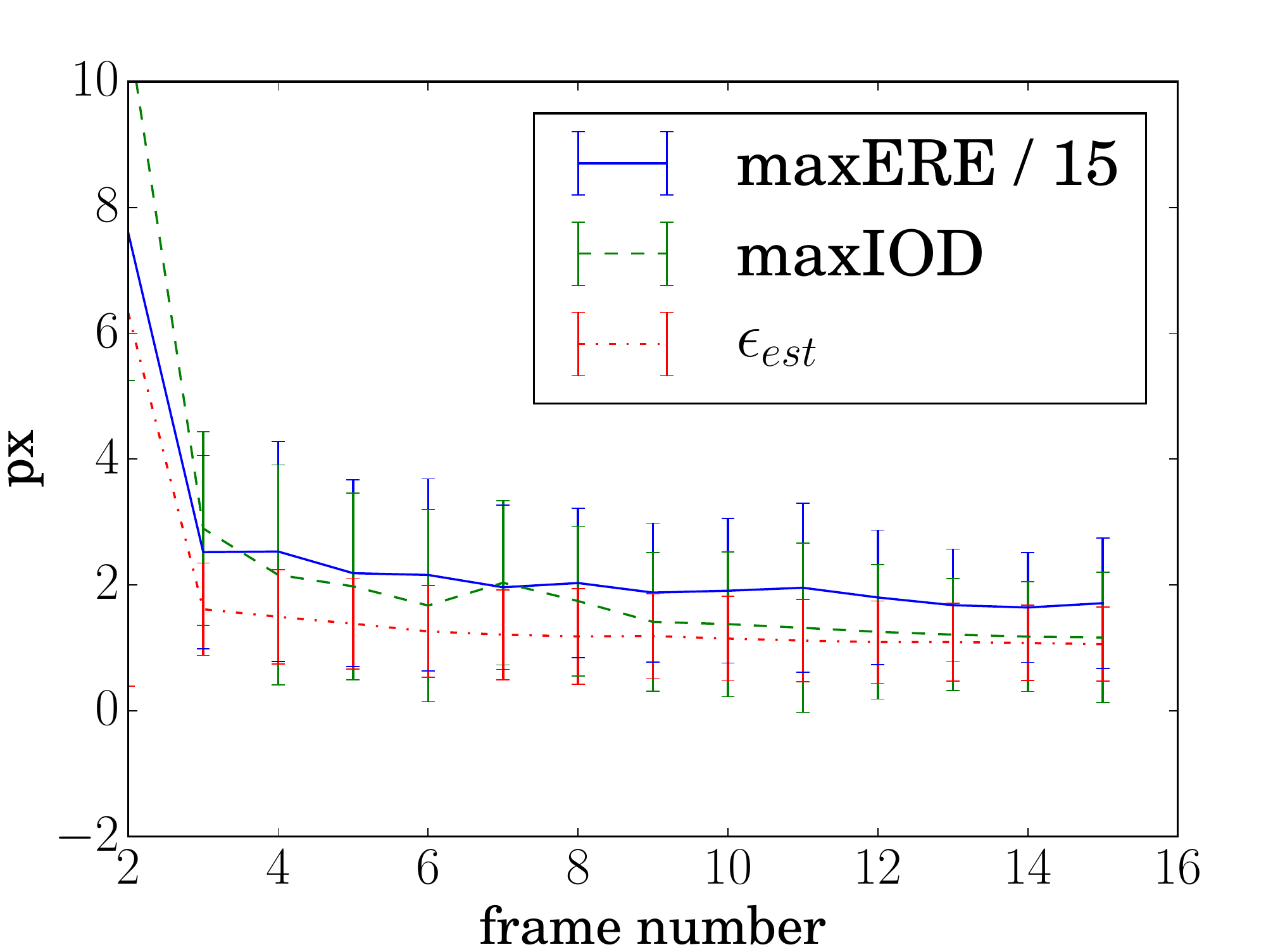}
\label{fig:error-metrics}
}
\subfloat[] {
\includegraphics[width=0.24\textwidth]{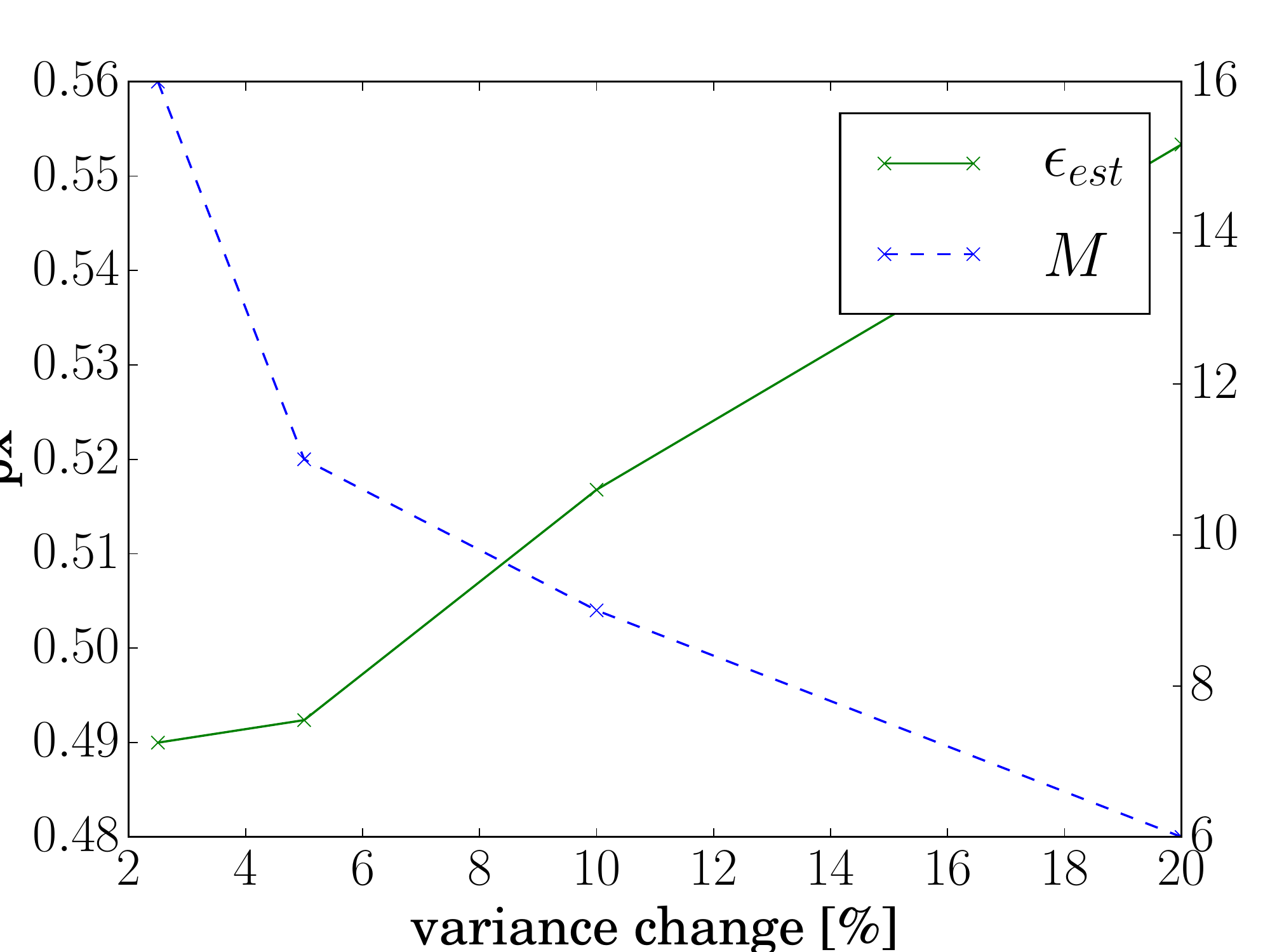}
\label{fig:quality-level}
}
\caption{\textit{(a)} Comparing error metrics on synthetic data: both MaxERE and the proposed Max IOD correlate with estimation error $\epsilon_{est}$. (standard deviation over 20 samples) \textit{(b)} Required number of frames $M$ and $\epsilon_{est}$ in respect to the variance reduction threshold}
\end{figure}

We also evaluated the usage of MaxIOD as an error metric by comparing it to MaxERE \cite{richardson2013iros} and a known estimation error $\epsilon_{est}$.
Just as the MaxERE, the MaxIOD correlates with $\epsilon_{est}$ (see Figure \ref{fig:error-metrics}). Additionally, as Figure \ref{fig:quality-level} indicates, the IOD reduction is suitable for balancing calibration quality and the number of required calibration frames.

\subsection{Real data}

For evaluating our method with real images, we recorded a separate testing set consisting of 50 images at various distances and angles covering the whole field of view.
All images were captured using a Logitech C525 webcam at a resolution of 1280x720px. The auto\-focus was fixed throughout the whole evaluation, while exposure was fixed per sequence.
Our method was compared to AprilCal \cite{richardson2013iros} and calibrating without any pose restrictions using OpenCV.

We used the pattern described in section \ref{sec:pattern} that provides 40 measurements per frame for OpenCV as well as for our method. With AprilCal, we used the 5x7 AprilTag target that generates approximately the same amount of measurements.

The convergence threshold was set to 10\% for our method and the stopping accuracy parameter of AprilCal was set to 2.0. As the OpenCV method does not provide convergence monitoring, we stopped calibration after 10 frames here.

\begin{table}
\centering
\begin{tabular}{|c|c|c|c|}
\hline 
Method & mean $\epsilon_{est}$ & frames used & mean $\epsilon_{res}$ \\ 
\hline 
\hline
Pose selection  & 0.518  & 9.4 & 0.470 \\ 
\hline 
OpenCV \cite{opencv_library}        & 1.465 & 10  & 0.345 \\
\hline
AprilCal \cite{richardson2013iros}  & 0.815 & 13.4 & 1.540 \\
\hline
Compactness test  & 0.514  & 7 & 0.476 \\ 
\hline 
\end{tabular} 
\caption{Our method compared to AprilCal and OpenCV on real data. Showing the average over five runs.
Training on the testing set results in $\epsilon_{est} = 0.479$.}
\label{tbl:evaluation}
\end{table}

Table \ref{tbl:evaluation} shows the mean results over 5 calibration runs for each method, measuring the required number of frames, $\epsilon_{est}$ and $\epsilon_{res}$.
Here our method requires only 70\% of the frames required by AprilCal while arriving at a 36\% lower $\epsilon_{est}$ (64\% compared to OpenCV).

\subsection{Analyzing the calibration compactness}
The results in the previous section show that our method is able to provide the lowest calibration error $\epsilon_{est}$ while using fewer calibration frames then comparable approaches.
However, it is not clear whether the solution is using the minimal amount of frames or whether it is possible to use a subset of frames while arriving at the same calibration error.

Therefore, we further tested the compactness of our calibration result. We used a greedy algorithm that, given a set of frames captured by our method, tries to find a smaller subset.
It optimizes for the testing set, directly minimizing the estimation error.

The algorithm is computed as follows; given a set of training images (the calibration sequence)
\begin{enumerate}[nolistsep]
\item the initialization frames as described in Section \ref{sec:init} are added unconditionally;
\item each of the remaining frames is now \textit{individually} added to the key-frame set and a calibration is computed.
\item For each calibration the estimation error $\epsilon_{est}$ is computed using the testing frames.
\item The frame that minimizes $\epsilon_{est}$ is incorporated into the key-frame set. Continue at step 2.
\item Terminate if $\epsilon_{est}$ cannot be further reduced or all frames have been used.
\end{enumerate}

The greedy optimal solution requires 75\% of the frames compared to the proposed method while keeping the same estimation error (see Table \ref{tbl:evaluation}).
This indicates that, while a significant improvement over \cite{richardson2013iros}, our method is not yet optimal in the compactness sense.
The greedy algorithm requires an a-priori recorded testing set and only finds a minimal subset of an existing calibration sequence, but cannot generate any calibration poses.

\subsection{User survey}

We performed an informal survey among 5 co-workers to measure the required calibration time when using our method. The tool was used for the first time and the only given instruction was that the overlay should be matched with the calibration pattern. The camera was fixed and the pattern had to be moved.
On average the users required 1:33 min for capturing 8.7 frames at $\epsilon_{est} = 0.533$.

\section{Conclusion and future work}
\label{sec:conclusion}

We have presented a calibration approach to generate a compact set of calibration frames that is suitable for interactive user guidance. Singular pose configurations are avoided such that capturing about 9 key-frames is sufficient for a precise calibration. This is 30\% less than comparable solutions.
The provided user guidance allows even inexperienced users to perform the calibration in less than 2 minutes. Calibration precision can be weighted against the required calibration time using the convergence threshold.
The camera parameter uncertainty is monitored throughout the processes, ensuring that a given confidence level can be reached repeatedly.

Our evaluation shows that the amount of required frames can still be reduced to speed up the process even more.
We only use a widespread and simple distortion model, additional distortion coefficients like thin prism \cite{weng1992camera}, rational \cite{ma2004rational} and tilted sensor are to be considered in future work. Eventually one could incorporate a detection of unused parameters. This would allow to start with the most complex distortion model which could be gradually reduced during calibration.

Furthermore the method needs adaptation to special cases like microscopy where the depth of field limits the possible calibration angles or calibration at large distance where scaling the pattern accordingly is not desirable.


The OpenCV based implementation of the presented algorithm is available open-source at \url{https://github.com/paroj/pose_calib}.

\bibliographystyle{abbrvnat}
\bibliography{bibliography}

\end{document}